\def\year{2018}\relax
\documentclass[letterpaper]{article} 
\usepackage{aaai18}  
\usepackage{times}  
\usepackage{helvet}  
\usepackage{courier}  
\usepackage{url}  
\usepackage{graphicx}  
\usepackage{algorithm}
\usepackage{algorithmicx}
\usepackage{algpseudocode}

\usepackage{latexsym}

\usepackage{amsmath,amssymb,amsthm}
\usepackage{multirow}
\usepackage{makecell}
\usepackage{url}
\usepackage{caption}
\usepackage{subcaption,siunitx,booktabs}
\usepackage{sidecap}
\usepackage{textcomp}
\usepackage{pbox}
\newcommand{\dynamicRNN}{\textsc{Dynamic-RNN}}
\newcommand{\staticRNN}{\textsc{Static-RNN}}
\newcommand{\staticHierRNN}{\textsc{Static-Hier-RNN}}
\newcommand{\recentTFIDF}{\textsc{Recent+TF-IDF}}
\newcommand{\directRecentTFIDF}{\textsc{Direct-Recent+TF-IDF}}

\newcommand{\newcite}[1]{\citeauthor{#1} \shortcite{#1}}

\newcommand{\newvec}[1]{\mathbf{#1}}

\DeclareMathOperator*{\argmax}{arg\,max}

\begin{document}
\title{Addressee and Response Selection in Multi-Party Conversations \\ with Speaker Interaction RNNs}
\author{Rui Zhang \\ Yale University \\ r.zhang@yale.edu \And Honglak Lee \\ University of Michigan \\ honglak@eecs.umich.edu \And Lazaros Polymenakos \\ IBM T. J. Watson Research Center \\ lcpolyme@us.ibm.com \And Dragomir Radev \\ Yale University \\ dragomir.radev@yale.edu}

\maketitle
\begin{abstract}
In this paper, we study the problem of addressee and response selection in multi-party conversations. 
Understanding multi-party conversations is challenging because of complex speaker interactions: multiple speakers exchange messages with each other, playing different roles (sender, addressee, observer), and these roles vary across turns.
To tackle this challenge, we propose the Speaker Interaction Recurrent Neural Network (SI-RNN).
Whereas the previous state-of-the-art system updated speaker embeddings only for the sender, SI-RNN uses a novel dialog encoder to update speaker embeddings in a role-sensitive way.
Additionally, unlike the previous work that selected the addressee and response separately, SI-RNN selects them jointly by viewing the task as a sequence prediction problem.
Experimental results show that SI-RNN significantly improves the accuracy of addressee and response selection, particularly in complex conversations with many speakers and responses to distant messages many turns in the past.
\end{abstract}
\section{Introduction}
Real-world conversations often involve more than two speakers.
In the Ubuntu Internet Relay Chat channel (IRC), for example, one user can initiate a discussion about an Ubuntu-related technical issue, and many other users can work together to solve the problem.
Dialogs can have complex speaker interactions: at each turn, users play one of three roles (sender, addressee, observer), and those roles vary across turns.

In this paper, we study the problem of addressee and response selection in multi-party conversations: given a responding speaker and a dialog context, the task is to select an addressee and a response from a set of candidates for the responding speaker.
The task requires modeling multi-party conversations and can be directly used to build retrieval-based dialog systems \cite{lu2013deep,hu2014convolutional,ji2014information,wang2015syntax}.

The previous state-of-the-art \dynamicRNN{} model from \newcite{ouchi-tsuboi:2016:EMNLP2016} maintains speaker embeddings to track each speaker status, which dynamically changes across time steps.
It then produces the context embedding from the speaker embeddings and selects the addressee and response based on embedding similarity.
However, this model updates only the sender embedding, not the embeddings of the addressee or observers, with the corresponding utterance, and it selects the addressee and response separately.
In this way, it only models \textit{who} says \textit{what} and fails to capture addressee information.
Experimental results show that the separate selection process often produces inconsistent addressee-response pairs.

To solve these issues, we introduce the Speaker Interaction Recurrent Neural Network (SI-RNN).
SI-RNN redesigns the dialog encoder by updating speaker embeddings in a role-sensitive way.
Speaker embeddings are updated in different GRU-based units depending on their roles (sender, addressee, observer).
Furthermore, we note that the addressee and response are mutually dependent and view the task as a joint prediction problem.
Therefore, SI-RNN models the conditional probability (of addressee given the response and vice versa) and selects the addressee and response pair by maximizing the joint probability.

On a public standard benchmark data set, SI-RNN significantly improves the addressee and response selection accuracy, particularly in complex conversations with many speakers and responses to distant messages many turns in the past.
Our code and data set are available online.\footnote{The released code: https://github.com/ryanzhumich/sirnn}

\section{Related Work}
We follow a data-driven approach to dialog systems.
\newcite{singh1999reinforcement}, \newcite{henderson2008hybrid}, and \newcite{young2013pomdp} optimize the dialog policy using Reinforcement Learning or the Partially Observable Markov Decision Process framework.
In addition, \newcite{henderson2014second} propose to use a predefined ontology as a logical representation for the information exchanged in the conversation.
The dialog system can be divided into different modules, such as Natural Language Understanding \cite{yao2014spoken,mesnil2015using}, Dialog State Tracking \cite{henderson2014word,williams2016dialog}, and Natural Language Generation \cite{wensclstm15}.
Furthermore, \newcite{wen2016network} and \newcite{bordes2017learning} propose end-to-end trainable goal-oriented dialog systems.

Recently, short text conversation has been popular.
The system receives a short dialog context and generates a response using statistical machine translation or sequence-to-sequence networks \cite{ritter2011data,vinyals2015neural,shang2015neural,serban2016building,li-EtAl:2016:P16-13,mei2017coherent}.
In contrast to response generation, the retrieval-based approach uses a ranking model to select the highest scoring response from candidates \cite{lu2013deep,hu2014convolutional,ji2014information,wang2015syntax}.
However, these models are single-turn responding machines and thus still are limited to short contexts with only two speakers.
As for larger context, \newcite{lowe2015ubuntu} propose the Next Utterance Classification (NUC) task for multi-turn two-party dialogs.
\newcite{ouchi-tsuboi:2016:EMNLP2016} extend NUC to multi-party conversations by integrating the addressee detection problem.
Since the data is text based, they use only textual information to predict addressees as opposed to relying on acoustic signals or gaze information in multimodal dialog systems \cite{jovanovic2006addressee,akker2009comparison}.

Furthermore, several other papers are recently presented focusing on modeling role-specific information given the dialogue contexts \cite{meng2017towards,chi2017speaker,chen2017dynamic}.
For example, \newcite{meng2017towards} combine content and temporal information to predict the utterance speaker.
By contrast, our SIRNN explicitly utilizes the speaker interaction to maintain speaker embeddings and predicts the addressee and response by joint selection.

\section{Preliminaries}
\subsection{Addressee and Response Selection}

\newcite{ouchi-tsuboi:2016:EMNLP2016} propose the addressee and response selection task for multi-party conversation.
Given a responding speaker $a_{res}$ and a dialog context $\mathcal{C}$, the task is to select a response and an addressee.
$\mathcal{C}$ is a list ordered by time step: 
$$ \mathcal{C} = [(a^{(t)}_{sender}, a^{(t)}_{addressee}, u^{(t)})]_{t=1}^T $$
where $a^{(t)}_{sender}$ says $u^{(t)}$ to $a^{(t)}_{addressee}$ at time step $t$,
and $T$ is the total number of time steps before the response and addressee selection.
The set of speakers appearing in $\mathcal{C}$ is denoted $\mathcal{A(C)}$.
As for the output, the addressee is selected from $\mathcal{A(C)}$, and the response is selected from a set of candidates $\mathcal{R}$.
Here, $\mathcal{R}$ contains the ground-truth response and one or more false responses.
%
We provide some examples in Table \ref{tb:example} (Section \ref{sec-results}).

\begin{table}[t]
\centering
\begin{tabular}{ccc}
                        & Data                & Notation \\ \Xhline{4\arrayrulewidth}
                        & Responding Speaker  & $a_{res}$         \\
Input                   & Context             & $\mathcal{C}$     \\
                        & Candidate Responses & $\mathcal{R}$     \\ \hline
\multirow{2}{*}{Output} & Addressee           & $a \in \mathcal{A(C)}$  \\
                        & Response            & $r \in \mathcal{R}$ \\ \hline
\multicolumn{2}{c}{Sender ID at time $t$} & $a^{(t)}_{sender}$\\
\multicolumn{2}{c}{Addressee ID at time $t$} & $a^{(t)}_{addressee}$ \\
\multicolumn{2}{c}{Utterance at time $t$} & $u^{(t)}$ \\ 
\multicolumn{2}{c}{Utterance embedding at time $t$} & $\newvec{u}^{(t)}$ \\ 
\multicolumn{2}{c}{Speaker embedding of $a_i$ at time $t$}  & $\newvec{a}_{i}^{(t)}$ \\
\end{tabular}
\caption{Notations for the task and model.}
\label{tab:notation}
\end{table}

\begin{figure*}[t!]
\centering
\begin{subfigure}{.5\textwidth}
  \centering
  \includegraphics[width=1.\textwidth]{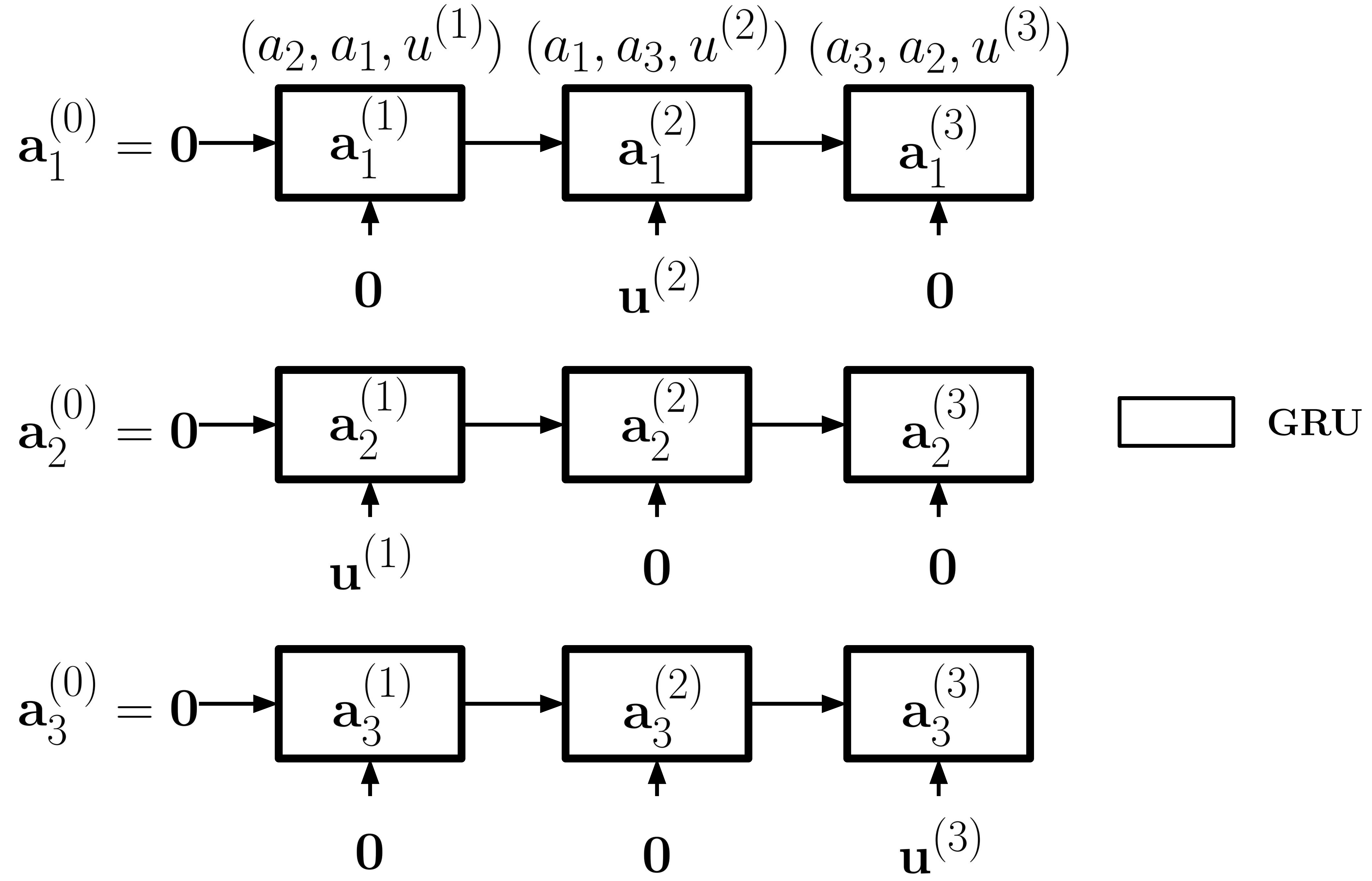}
\end{subfigure}%
\begin{subfigure}{.5\textwidth}
  \centering
  \includegraphics[width=1.\textwidth]{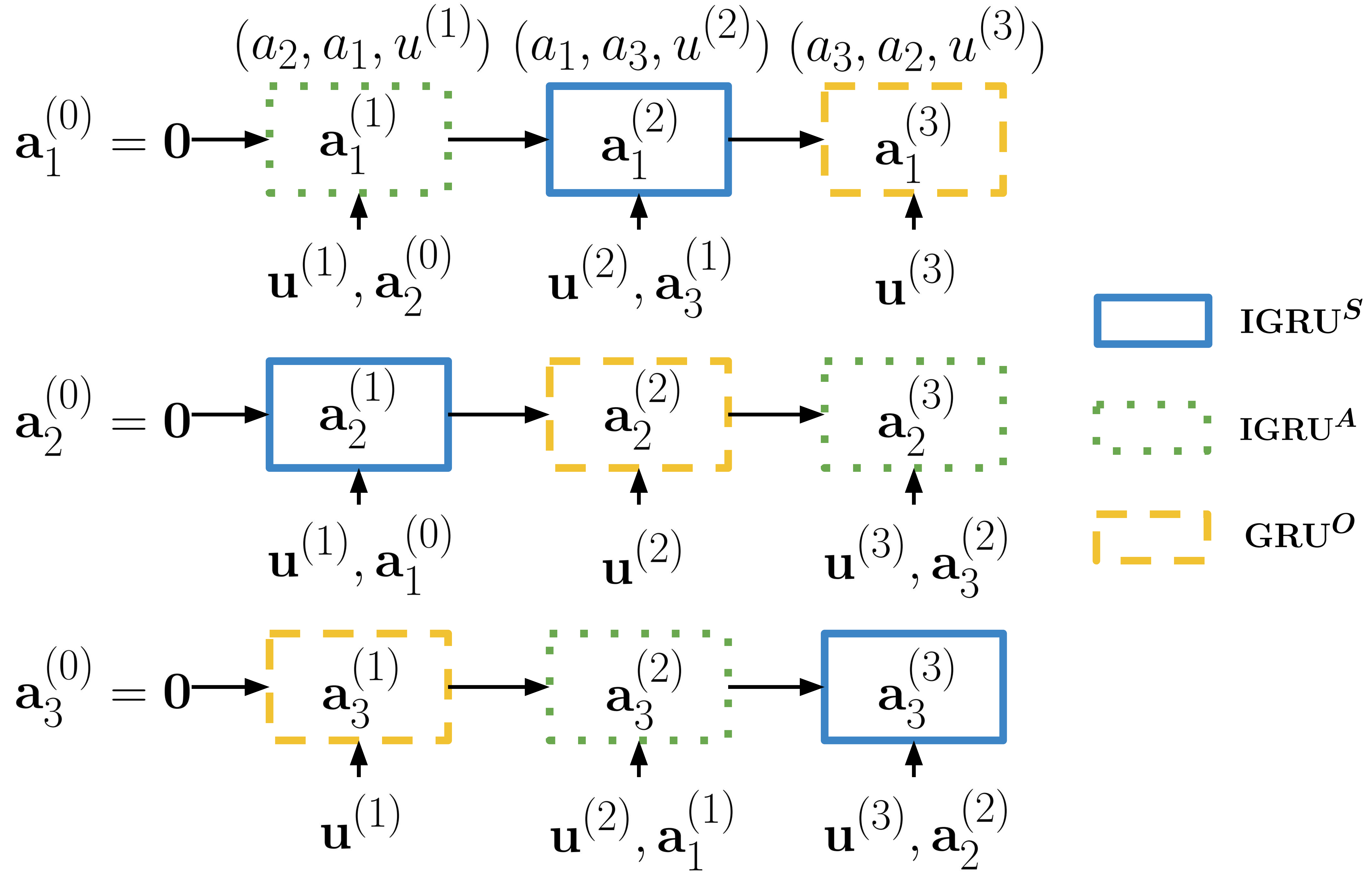}
\end{subfigure}
\caption{Dialog encoders in \dynamicRNN{} (Left) and SI-RNN (Right) for an example context at the top.
Speaker embeddings are initialized as zero vectors and updated recurrently as hidden states along the time step. In SI-RNN, the same speaker embedding is updated in different units depending on the role ($\mathrm{IGRU}^{S}$ for sender, $\mathrm{IGRU}^{A}$ for addressee, $\mathrm{GRU}^{O}$ for observer).}
\label{fig:dynamic-sirnn}
\end{figure*}

\subsection{\dynamicRNN{} Model}
\label{sec:dynamic}
In this section, we briefly review the state-of-the-art \dynamicRNN{} model \cite{ouchi-tsuboi:2016:EMNLP2016}, which our proposed model is based on.
\dynamicRNN{} solves the task in two phases:
1) the dialog encoder maintains a set of speaker embeddings to track each speaker status, which dynamically changes with time step $t$;
2) then \dynamicRNN{} produces the context embedding from the speaker embeddings and selects the addressee and response based on embedding similarity among context, speaker, and utterance.

\paragraph{Dialog Encoder.}
Figure \ref{fig:dynamic-sirnn} (Left) illustrates the dialog encoder in \dynamicRNN{} on an example context.
In this example, $a_2$ says $u^{(1)}$ to $a_1$, then $a_1$ says $u^{(2)}$ to $a_3$, and finally $a_3$ says $u^{(3)}$ to $a_2$.
The context $\mathcal{C}$ will be:
\begin{equation}
\label{eq:context}
\!\!\!\!\!\!    \mathcal{C} = [(a_2,a_1,u^{(1)}), (a_1,a_3,u^{(2)}), (a_3,a_2,u^{(3)})]
\end{equation}
with the set of speakers $\mathcal{A(C)} = \{a_1,a_2,a_3\}$.

For a speaker $a_i$, the bold letter $\newvec{a}_{i}^{(t)} \in \mathbb{R}^{d_s}$ denotes its embedding at time step $t$.
Speaker embeddings are initialized as zero vectors and updated recurrently as hidden states of GRUs \cite{cho-EtAl:2014:EMNLP2014,chung2014empirical}.
Specifically, for each time step $t$ with the sender $a_{sdr}$ and the utterance $u^{(t)}$, the sender embedding $\newvec{a}_{sdr}$ is updated recurrently from the utterance:
$$\newvec{a}_{sdr}^{(t)} = \mathrm{GRU}(\newvec{a}_{sdr}^{(t-1)},\newvec{u}^{(t)}),$$
where $\newvec{u}^{(t)} \in \mathbb{R}^{d_u}$ is the embedding for utterance $u^{(t)}$.
Other speaker embeddings are updated from $\newvec{u}^{(t)} = \newvec{0}$.
The speaker embeddings are updated until time step $T$.

\paragraph{Selection Model.}
To summarize the whole dialog context $\mathcal{C}$, the model applies element-wise max pooling over all the speaker embeddings to get the context embedding $\newvec{h}_{\mathcal{C}}$:
\begin{equation}
\label{eq:maxpool}
\newvec{h}_{\mathcal{C}} = \max_{a_i=a_1,\dots,a_{|\mathcal{A(C)}|}} \newvec{a}_{i}^{(T)} \in \mathbb{R}^{d_s}
\end{equation}

The probability of an addressee and a response being the ground truth is calculated based on embedding similarity.
To be specific, for addressee selection, the model compares the candidate speaker $a_p$, the dialog context $\mathcal{C}$, and the responding speaker $a_{res}$:
\begin{equation}
\label{eq:pa}
\mathbb{P}(a_p|\mathcal{C}) = \sigma([\newvec{a}_{res};\newvec{h}_{\mathcal{C}}]^\top \newvec{W}_a \newvec{a}_{p})
\end{equation}
where $\newvec{a}_{res}$ is the final speaker embedding for the responding speaker $a_{res}$, $\newvec{a}_{p}$ is the final speaker embedding for the candidate addressee $a_p$, 
$\sigma$ is the logistic sigmoid function, $[\, ; \, ]$ is the row-wise concatenation operator, and $\newvec{W}_a \in \mathbb{R}^{2d_s \times d_s}$ is a learnable parameter.
Similarly, for response selection, 
\begin{equation}
\label{eq:pr}
\mathbb{P}(r_q|\mathcal{C}) = \sigma([\newvec{a}_{res};\newvec{h}_{\mathcal{C}}]^\top \newvec{W}_r \newvec{r}_{q})
\end{equation}
where $\newvec{r}_{q} \in \mathbb{R}^{d_u}$ is the embedding for the candidate response $r_q$, and $\newvec{W}_r \in \mathbb{R}^{2d_s \times d_u}$ is a learnable parameter.

The model is trained end-to-end to minimize a joint cross-entropy loss for the addressee selection and the response selection with equal weights. 
At test time, the addressee and the response are separately selected to maximize the probability in Eq \ref{eq:pa} and Eq \ref{eq:pr}.

\section{Speaker Interaction RNN}
While \dynamicRNN{} can track the speaker status by capturing \textit{who} says \textit{what} in multi-party conversation, there are still some issues.
First, at each time step, only the sender embedding is updated from the utterance.
Therefore, other speakers are blind to what is being said,
and the model fails to capture addressee information.
Second, while the addressee and response are mutually dependent, \dynamicRNN{} selects them independently.
Consider a case where the responding speaker is talking to two other speakers in separate conversation threads.
The choice of addressee is likely to be either of the two speakers, but the choice is much less ambiguous if the correct response is given, and vice versa.
\dynamicRNN{} often produces inconsistent addressee-response pairs due to the separate selection.
See Table \ref{tb:example} for examples.

In contrast to \dynamicRNN, the dialog encoder in SI-RNN updates embeddings for all the speakers besides the sender at each time step.
Speaker embeddings are updated depending on their roles: the update of the sender is different from the addressee, which is different from the observers.
Furthermore, the update of a speaker embedding is not only from the utterance, but also from other speakers.
These are achieved by designing variations of GRUs for different roles.
Finally, SI-RNN selects the addressee and response jointly by maximizing the joint probability.

\begin{algorithm}[t]
\caption{Dialog Encoder in SI-RNN}
\begin{algorithmic}[1]
\State \textbf{Input} 
\State $\textrm{Dialog context } \mathcal{C} \textrm{ with speakers } \mathcal{A(C)}$:
\State $\mathcal{C} = [(a^{(t)}_{sender}, a^{(t)}_{addressee}, u^{(t)} )]_{t=1}^T$ 
\State $\mathcal{A(C)} = \{a_1,a_2,\dots,a_{|\mathcal{A(C)}|}\}$
\State where $a^{(t)}_{sender}, a^{(t)}_{addressee} \in \mathcal{A(C)}$
\State // Initialize speaker embeddings
\For{$a_i=a_1,a_2,\dots,a_{|\mathcal{A(C)}|}$}
\State $\newvec{a}^{(0)}_{i} \gets \newvec{0} \in \mathbb{R}^{d_s}$
\EndFor
\State //Update speaker embeddings
\For{$t = 1,2,\dots,T$}
  \State // Update sender, addressee, observers
  \State $a_{sdr} \gets a^{(t)}_{sender}$
  \State $a_{adr} \gets a^{(t)}_{addressee}$
  \State $\mathcal{O} \gets  \mathcal{A(C)} - \{a_{sdr}, a_{adr} \}$
  \State // Compute utterance embedding
  \State $\newvec{u}^{(t)} \gets \mathrm{UtteranceEncoder}(u^{(t)})$
  \State $\newvec{u}^{(t)} \gets \mathrm{Concatenate}(\newvec{a}_{sdr}^{(t-1)},\newvec{u}^{(t)})$
  \State // Update sender embedding
  \State $\newvec{a}_{sdr}^{(t)} \gets \mathrm{IGRU}^{S}(\newvec{a}_{sdr}^{(t-1)},\newvec{a}_{adr}^{(t-1)},\newvec{u}^{(t)})$
  \State // Update addressee embedding
  \State $\newvec{a}_{adr}^{(t)} \gets \mathrm{IGRU}^{A}(\newvec{a}_{adr}^{(t-1)},\newvec{a}_{sdr}^{(t-1)},\newvec{u}^{(t)})$
  \State // Update observer embeddings
  \For{$a_{obr} \textbf{ in } \mathcal{O}$}
    \State $\newvec{a}_{obr}^{(t)} \gets \mathrm{GRU}^{O}(\newvec{a}_{obr}^{(t-1)},\newvec{u}^{(t)})$
  \EndFor
\EndFor
\State // Return final speaker embeddings
\State \textbf{Output} 
\State \Return $\newvec{a}_{i}^{(T)}$ for $a_i=a_1,a_2,\dots,a_{|\mathcal{A(C)}|}$
\end{algorithmic}
\label{algo:dialog_encoder}
\end{algorithm}

\subsection{Utterance Encoder}
\label{sec:utterance_encoder}
To encode an utterance $u = (w_1, w_2,..., w_N)$ of $N$ words, we use a RNN with Gated Recurrent Units \cite{cho-EtAl:2014:EMNLP2014,chung2014empirical}:
$$\newvec{h}_{j} = \mathrm{GRU}(\newvec{h}_{j-1},\newvec{x}_{j})$$
where $\newvec{x}_j$ is the word embedding for $w_j$, and $\newvec{h}_j$ is the $\mathrm{GRU}$ hidden state.
$\newvec{h}_0$ is initialized as a zero vector, and the utterance embedding is the last hidden state, i.e. $\newvec{u} = \newvec{h}_N$.

\subsection{Dialog Encoder}
\label{sec:dialog_encoder}

Figure \ref{fig:dynamic-sirnn} (Right) shows how SI-RNN encodes the example in Eq \ref{eq:context}.
Unlike \dynamicRNN, SI-RNN updates all speaker embeddings in a role-sensitive manner.
For example, at the first time step when $a_2$ says $u^{(1)}$ to $a_1$, \dynamicRNN{} only updates $\newvec{a}_2$ using $\newvec{u}^{(1)}$, while other speakers are updated using $\newvec{0}$.
In contrast, SI-RNN updates each speaker status with different units: $\mathrm{IGRU}^{S}$ updates the sender embedding $\newvec{a}_2$ from the utterance embedding $\newvec{u}^{(1)}$ and the addressee embedding $\newvec{a}_1$; $\mathrm{IGRU}^{A}$ updates the addressee embedding $\newvec{a}_1$ from $\newvec{u}^{(1)}$ and $\newvec{a}_2$; $\mathrm{GRU}^{O}$ updates the observer embedding $\newvec{a}_3$ from $\newvec{u}^{(1)}$.

Algorithm \ref{algo:dialog_encoder} gives a formal definition of the dialog encoder in SI-RNN.
The dialog encoder is a function that takes as input a dialog context $\mathcal{C}$ (lines 1-5) and returns speaker embeddings at the final time step (lines 28-30).
Speaker embeddings are initialized as $d_s$-dimensional zero vectors (lines 6-9).
Speaker embeddings are updated by iterating over each line in the context (lines 10-27).

\subsection{Role-Sensitive Update}
\begin{figure}[t]
  \centering
  \includegraphics[width=0.33\textwidth]{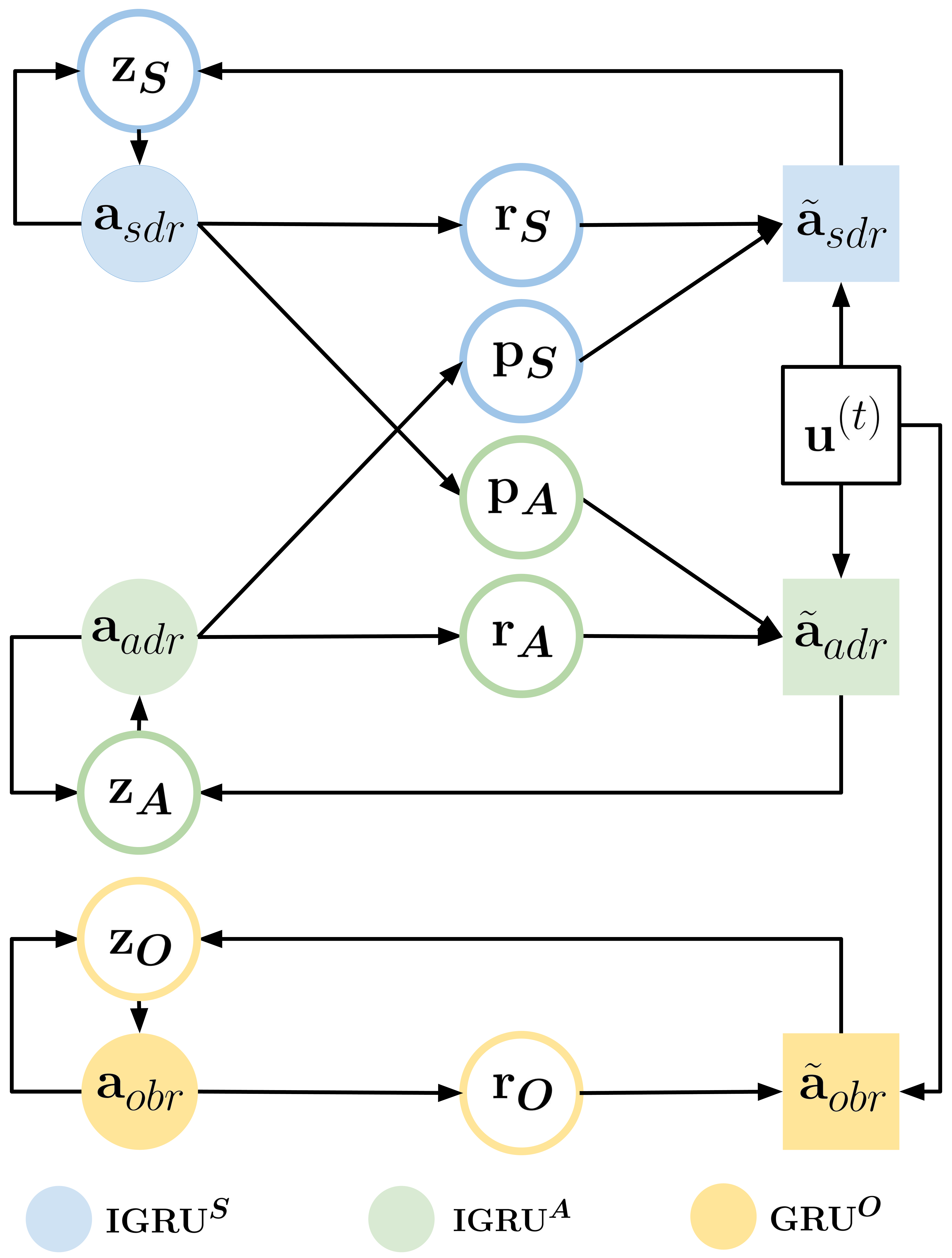}
  \caption{Illustration of $\mathrm{IGRU}^{S}$ (upper, blue), $\mathrm{IGRU}^{A}$ (middle, green), and $\mathrm{GRU}^{O}$ (lower, yellow).
  Filled circles are speaker embeddings, which are recurrently updated. Unfilled circles are gates. Filled squares are speaker embedding proposals.
  }
  \label{fig:igru}
\end{figure}

In this subsection, we explain in detail how $\mathrm{IGRU}^{S}$/$\mathrm{IGRU}^{A}$/$\mathrm{GRU}^{O}$ update speaker embeddings according to their roles at each time step (Algorithm \ref{algo:dialog_encoder} lines 19-26).

As shown in Figure \ref{fig:igru}, $\mathrm{IGRU}^{S}$/$\mathrm{IGRU}^{A}$/$\mathrm{GRU}^{O}$ are all GRU-based units.
$\mathrm{IGRU}^{S}$ updates the sender embedding from the previous sender embedding $\newvec{a}_{sdr}^{(t-1)}$, the previous addressee embedding $\newvec{a}_{adr}^{(t-1)}$, and the utterance embedding $\newvec{u}^{(t)}$:
$$\newvec{a}_{sdr}^{(t)} \gets \mathrm{IGRU}^{S}(\newvec{a}_{sdr}^{(t-1)},\newvec{a}_{adr}^{(t-1)},\newvec{u}^{(t)})$$
The update, as illustrated in the upper part of Figure \ref{fig:igru}, is controlled by three gates.
The $\newvec{r}^{(t)}_S$ gate controls the previous sender embedding $\newvec{a}_{sdr}^{(t-1)}$, and $\newvec{p}^{(t)}_S$ controls the previous addressee embedding $\newvec{a}_{adr}^{(t-1)}$.
Those two gated interactions together produce the sender embedding proposal $\newvec{\widetilde{a}}_{sdr}^{(t)}$.
Finally, the update gate $\newvec{z}^{(t)}_S$ combines the proposal $\newvec{\widetilde{a}}_{sdr}^{(t)}$ and the previous sender embedding $\newvec{a}_{sdr}^{(t-1)}$ to update the sender embedding $\newvec{a}_{sdr}^{(t)}$.
The computations in $\mathrm{IGRU}^{S}$ (including gates $\newvec{r}^{(t)}_S$, $\newvec{p}^{(t)}_S$, $\newvec{z}^{(t)}_S$, the proposal embedding $\newvec{\widetilde{a}}_{sdr}^{(t)}$, and the final updated embedding $\newvec{a}_{sdr}^{(t)}$) are formulated as:
\begin{align*}
\begin{split}
\newvec{r}^{(t)}_S =& \sigma( \newvec{W}_S^{r}\newvec{u}^{(t)} + \newvec{U}_S^{r}\newvec{a}_{sdr}^{(t-1)} + \newvec{V}_S^{r}\newvec{a}_{adr}^{(t-1)} )\\
\newvec{p}^{(t)}_S =& \sigma( \newvec{W}_S^{p}\newvec{u}^{(t)} + \newvec{U}_S^{p}\newvec{a}_{sdr}^{(t-1)} + \newvec{V}_S^{p}\newvec{a}_{adr}^{(t-1)} )\\
\newvec{z}^{(t)}_S =& \sigma( \newvec{W}_S^{z}\newvec{u}^{(t)} + \newvec{U}_S^{z}\newvec{a}_{sdr}^{(t-1)} + \newvec{V}_S^{z}\newvec{a}_{adr}^{(t-1)} )\\
\newvec{\widetilde{a}}_{sdr}^{(t)} = &\tanh(\newvec{W}_S\newvec{u}^{(t)} + \newvec{U}_S(\newvec{r}^{(t)}_S \odot \newvec{a}_{sdr}^{(t-1)}) \\
                        & \qquad + \newvec{V}_S(\newvec{p}^{(t)}_S \odot \newvec{a}_{adr}^{(t-1)}))\\
\newvec{a}_{sdr}^{(t)} &= \newvec{z}^{(t)}_S \odot \newvec{a}_{sdr}^{(t-1)} + (1-\newvec{z}^{(t)}_S) \odot \newvec{\widetilde{a}}_{sdr}^{(t)}\\
\end{split}
\end{align*}
where $\{\newvec{W}_S^{r},\newvec{W}_S^{p},\newvec{W}_S^{z},\newvec{U}_S^{r},\newvec{U}_S^{p},\newvec{U}_S^{z},\newvec{V}_S^{r},\newvec{V}_S^{p},\newvec{V}_S^{z}, \newvec{W}_S,$
$\newvec{U}_S,\newvec{V}_S\}$ are learnable parameters.
$\mathrm{IGRU}^{A}$ uses the same formulation with a different set of parameters, as illustrated in the middle of Figure \ref{fig:igru}.
In addition, we update the observer embeddings from the utterance.
$\mathrm{GRU}^{O}$ is implemented as the traditional GRU unit in the lower part of Figure \ref{fig:igru}.
Note that the parameters in $\mathrm{IGRU}^{S}$/$\mathrm{IGRU}^{A}$/$\mathrm{GRU}^{O}$ are not shared.
This allows SI-RNN to learn role-dependent features to control speaker embedding updates.
The formulations of $\mathrm{IGRU}^{A}$ and $\mathrm{GRU}^{O}$ are similar.

\subsection{Joint Selection}
\label{sec:selection}
The dialog encoder takes the dialog context $\mathcal{C}$ as input and returns speaker embeddings at the final time step, $\newvec{a}_{i}^{(T)}$.
Recall from Section \ref{sec:dynamic} that \dynamicRNN{} produces the context embedding $\newvec{h}_{\mathcal{C}}$ using Eq \ref{eq:maxpool} and then selects the addressee and response separately using Eq \ref{eq:pa} and Eq \ref{eq:pr}.

In contrast, SI-RNN performs addressee and response selection jointly: the response is dependent on the addressee and vice versa.
Therefore, we view the task as a sequence prediction process:
given the context and responding speaker, we first predict the addressee, and then predict the response given the addressee.
(We also use the reversed prediction order as in Eq \ref{eq:joint}.)

In addition to Eq \ref{eq:pa} and Eq \ref{eq:pr}, SI-RNN is also trained to model the conditional probability as follows.
To predict the addressee, we calculate the probability of the candidate speaker $a_p$ to be the ground-truth \textit{given} the ground-truth response $r$ (available during training time):
\begin{equation}
\label{eq:par}
\mathbb{P}(a_p|\mathcal{C},r) = \sigma([\newvec{a}_{res};\newvec{h}_{\mathcal{C}};\newvec{r}]^\top \newvec{W}_{ar} \newvec{a}_{p})
\end{equation}
The key difference from Eq \ref{eq:pa} is that Eq \ref{eq:par} is conditioned on the correct response $r$ with embedding $\newvec{r}$.
Similarly, for response selection, we calculate the probability of a candidate response $r_q$ \textit{given} the ground-truth addressee $a_{adr}$:
\begin{equation}
\label{eq:pra}
\mathbb{P}(r_q|\mathcal{C},a_{adr}) = \sigma([\newvec{a}_{res};\newvec{h}_{\mathcal{C}};\newvec{a}_{adr}]^\top \newvec{W}_{ra} \newvec{r}_{q})
\end{equation}


At test time, SI-RNN selects the addressee-response pair from $\mathcal{A(C)} \times \mathcal{R}$ to maximize the joint probability $\mathbb{P}(r_q,a_p|\mathcal{C})$:
\begin{align}
\label{eq:joint}
\begin{split}
\hat{a},\hat{r} = \argmax_{a_p,r_q \in \mathcal{A(C)} \times \mathcal{R}} & \mathbb{P}(r_q,a_p|\mathcal{C}) \\
 = \argmax_{a_p,r_q \in \mathcal{A(C)} \times \mathcal{R}} & \mathbb{P}(r_q|\mathcal{C}) \cdot \mathbb{P}(a_p|\mathcal{C},r_q) \\ 
                &+\mathbb{P}(a_p|\mathcal{C}) \cdot \mathbb{P}(r_q|\mathcal{C},a_p) 
\end{split}
\end{align}
In Eq \ref{eq:joint}, we decompose the joint probability into two terms: the first term selects the response given the context, and then selects the addressee given the context and the selected response; the second term selects the addressee and response in the reversed order.\footnote{Detail: We also considered an alternative decomposition of the joint probability as $\log\mathbb{P}(r_{q},a_{p}|\mathcal{C})=\frac{1}{2} [\log\mathbb{P}(r_{q}|\mathcal{C})+\log\mathbb{P}(a_{p}|\mathcal{C},r_{q})+\log\mathbb{P}(a_{p}|\mathcal{C})+\log\mathbb{P}(r_{q}|\mathcal{C},a_{p})]$, but the performance was similar to Eq~\ref{eq:joint}.}

\begin{table*}[ht!]
\centering
\small
\begin{tabular}{c|c|cccc|cccc}
                           &    & \multicolumn{4}{c|}{RES-CAND = 2} & \multicolumn{4}{c}{RES-CAND = 10} \\ \cline{3-10}
                           &    & \multicolumn{1}{c|}{DEV} & \multicolumn{3}{c|}{TEST} & \multicolumn{1}{c|}{DEV} & \multicolumn{3}{c}{TEST} \\ \cline{3-10}
                           & $T$ & \multicolumn{1}{c|}{ADR-RES} & ADR-RES  & ADR     & RES    & \multicolumn{1}{c|}{ADR-RES} & ADR-RES & ADR   & RES     \\ \Xhline{4\arrayrulewidth}
Chance                       & -  & 0.62 & 0.62     & 1.24    & 50.00  & 0.12 & 0.12    & 1.24  & 10.00   \\
Recent+TF-IDF              & 15 & 37.11 & 37.13    & 55.62   & 67.89   & 14.91 & 15.44   & 55.62 & 29.19   \\
Direct-Recent+TF-IDF       & 15 & 45.83 & 45.76    & 67.72   & 67.89   & 18.94 & 19.50 & 67.72 & 29.40   \\ \hline 
Static-RNN{}       & 5  & 47.08 & 46.99    & 60.39   & 75.07   & 21.96 & 21.98   & 60.26 & 33.27   \\
\cite{ouchi-tsuboi:2016:EMNLP2016} & 10 & 48.52 & 48.67    & 60.97   & 77.75  & 22.78 & 23.31   & 60.66 & 35.91   \\
                             & 15 & 49.03 & 49.27    & 61.95   & 78.14   & 23.73 & 23.49   & 60.98 & 36.58   \\ \hline
Static-Hier-RNN               & 5  & 49.19 & 49.38    & 62.20   & 76.70   & 23.68 & 23.75 & 62.24 & 34.51 \\
\cite{zhou16multi}        & 10 & 51.37 & 51.76    & 64.61   & 78.28   & 25.46 & 25.83 & 64.86 & 36.94 \\
\cite{serban2016building} & 15 & 52.78 & 53.04    & 65.84   & 79.08   & 26.31 & 26.62 & 65.89 & 37.85 \\ \hline
Dynamic-RNN  & 5  & 49.38 & 49.80    & 63.19   & 76.07   & 23.44 & 23.72 & 63.28 & 33.62 \\
\cite{ouchi-tsuboi:2016:EMNLP2016} & 10 & 52.76 & 53.85    & 66.94   & 78.16   & 25.44 & 25.95 & 66.70 & 36.14 \\
& 15 & 54.45 & 54.88    & 68.54   & 78.64   & 26.73 & 27.19 & 68.41 & 36.93 \\ \hline
               & 5 & 60.57 & 60.69 & 74.08 & 78.14 & 30.65 & 30.71 & 72.59 & 36.45 \\ 
SI-RNN (Ours)  & 10 & 65.34 & 65.63 & 78.76 & 80.34 & 34.18 & 34.09 & 77.13 & 39.20 \\  
  & 15 & \textbf{67.01} & \textbf{67.30} & \textbf{80.47} & \textbf{80.91} & \textbf{35.50} & \textbf{35.76} & \textbf{78.53} & \textbf{40.83} \\
\hline
SI-RNN w/ shared IGRUs        & 15 & 59.50 & 59.47 & 74.20 & 78.08 & 28.31 & 28.45 & 73.35 & 36.00 \\
SI-RNN w/o joint selection  & 15 & 63.13  & 63.40 & 77.56 & 80.38 & 32.24 & 32.53 & 77.61 & 39.73 \\
\end{tabular}
\caption{Addressee and response selection results on the Ubuntu Multiparty Conversation Corpus. Metrics include accuracy of addressee selection (\texttt{ADR}), response selection (\texttt{RES}), and pair selection (\texttt{ADR-RES}). \texttt{RES-CAND}: the number of candidate responses. $T$: the context length.}
\label{tab:ubuntu_result}
\end{table*}

\begin{table}[t]
\centering
\small
\begin{tabular}{l|cccc}
                      & Total  & Train  & Dev    & Test   \\ \Xhline{4\arrayrulewidth}
\# Docs           & 7,355  & 6,606  & 367    & 382  \\
\# Utters         & 2.4M   & 2.1M   & 132.4k & 151.3k \\
\# Samples        & -      & 665.6k & 45.1k  & 51.9k \\
Adr Mention Freq & -      & 0.32   & 0.34   & 0.34 \\
\# Speakers / Doc   & 26.8   & 26.3   & 30.7   & 32.1 \\
\# Utters / Doc     & 326.3  & 317.9  & 360.8  & 396.1 \\
\# Words / Utter    & 11.1   & 11.1   & 11.2   & 11.3 \\
\end{tabular}
\caption{Data Statistics. ``Adr Mention Freq" is the frequency of explicit addressee mention.}
\label{tab:ubuntu_data}
\end{table}

\section{Experimental Setup}
\textbf{Data Set.}
We use the Ubuntu Multiparty Conversation Corpus \cite{ouchi-tsuboi:2016:EMNLP2016} and summarize the data statistics in Table \ref{tab:ubuntu_data}.
\urlstyle{same}
The whole data set (including the Train/Dev/Test split and the false response candidates) is publicly available.\footnote{\url{https://github.com/hiroki13/response-ranking/tree/master/data/input}} 
The data set is built from the Ubuntu IRC chat room where a number of users discuss Ubuntu-related technical issues.
The log is organized as one file per day corresponding to a document $\mathcal{D}$.
Each document consists of (Time, SenderID, Utterance) lines.
If users explicitly mention addressees at the beginning of the utterance, the addresseeID is extracted.
Then a sample, namely a unit of input (the dialog context and the current sender) and output (the addressee and response prediction) for the task, is created to predict the ground-truth addressee and response of this line.
Note that samples are created only when the addressee is explicitly mentioned for clear, unambiguous ground-truth labels.
False response candidates are randomly chosen from all other utterances \textit{within the same document}.
Therefore, distractors are likely from the same sub-conversation or even from the same sender but at different time steps.
This makes it harder than \newcite{lowe2015ubuntu} where distractors are randomly chosen from all documents.
If no addressee is explicitly mentioned, the addressee is left blank and the line is marked as a part of the context.
\\\textbf{Baselines.}
Apart from \dynamicRNN{}, we also include several other baselines.
\recentTFIDF{} always selects the most recent speaker (except the responding speaker $a_{res}$) as the addressee and chooses the response to maximize the tf-idf cosine similarity with the context.
We improve it by using a slightly different addressee selection heuristic (\directRecentTFIDF): select the most recent speaker that \textit{directly talks to} $a_{res}$ by an explicit addressee mention.
We select from the previous 15 utterances, which is the longest context among all the experiments.
This works much better when there are multiple concurrent sub-conversations, and $a_{res}$ responds to a distant message in the context.
We also include another GRU-based model \staticRNN{} from \newcite{ouchi-tsuboi:2016:EMNLP2016}.
Unlike \dynamicRNN, speaker embeddings in \staticRNN{} are based on the order of speakers and are fixed.
Furthermore, inspired by \newcite{zhou16multi} and \newcite{serban2016building}, we implement \staticHierRNN, a hierarchical version of \staticRNN.
It first builds utterance embeddings from words and then uses high-level RNNs to process utterance embeddings.
\\\textbf{Implementation Details}
For a fair comparison, we follow the hyperparameters from \newcite{ouchi-tsuboi:2016:EMNLP2016}, which are chosen based on the validation data set.
We take a maximum of 20 words for each utterance.
We use 300-dimensional GloVe word vectors\footnote{\url{http://nlp.stanford.edu/projects/glove/}}, which are fixed during training.
SI-RNN uses 50-dimensional vectors for both speaker embeddings and hidden states.
Model parameters are initialized with a uniform distribution between -0.01 and 0.01.
We set the mini-batch size to 128.
The joint cross-entropy loss function with 0.001 L2 weight decay is minimized by Adam \cite{kingma2014adam}.
The training is stopped early if the validation accuracy is not improved for 5 consecutive epochs.
All experiments are performed on a single GTX Titan X GPU.
The maximum number of epochs is 30, and most models converge within 10 epochs.

\section{Results and Discussion}
\label{sec-results}
For fair and meaningful quantitative comparisons, we follow \newcite{ouchi-tsuboi:2016:EMNLP2016}'s evaluation protocols.
SI-RNN improves the overall accuracy on the addressee and response selection task.
Two ablation experiments further analyze the contribution of role-sensitive units and joint selection respectively. 
We then confirm the robustness of SI-RNN with the number of speakers and distant responses.
Finally, in a case study we discuss how SI-RNN handles complex conversations by either engaging in a new sub-conversation or responding to a distant message.
\\\textbf{Overall Result.}
As shown in Table \ref{tab:ubuntu_result}, SI-RNN significantly improves upon the previous state-of-the-art.
In particular, addressee selection (\texttt{ADR}) benefits most, with different number of candidate responses (denoted as \texttt{RES-CAND}): around 12\% in
\texttt{RES-CAND} $= 2$ and more than 10\% in \texttt{RES-CAND} $= 10$.
Response selection (\texttt{RES}) is also improved, suggesting role-sensitive GRUs and joint selection are helpful for response selection as well.
The improvement is more obvious with more candidate responses (2\% in \texttt{RES-CAND} $= 2$ and 4\% in \texttt{RES-CAND} $= 10$).
These together result in significantly better accuracy on the \texttt{ADR-RES} metric as well.
\\\textbf{Ablation Study.}
We show an ablation study in the last rows of Table \ref{tab:ubuntu_result}.
First, we share the parameters of $\mathrm{IGRU}^{S}$/$\mathrm{IGRU}^{A}$/$\mathrm{GRU}^{O}$.
The accuracy decreases significantly, indicating that it is crucial to learn role-sensitive units to update speaker embeddings.
Second, to examine our joint selection, we fall back to selecting the addressee and response separately, as in \dynamicRNN.
We find that joint selection improves \texttt{ADR} and \texttt{RES} individually, and it is particularly helpful for pair selection \texttt{ADR-RES}.

\begin{figure}[t!]
\centering
\begin{subfigure}{.45\textwidth}
  \centering
  \includegraphics[width=.8\textwidth]{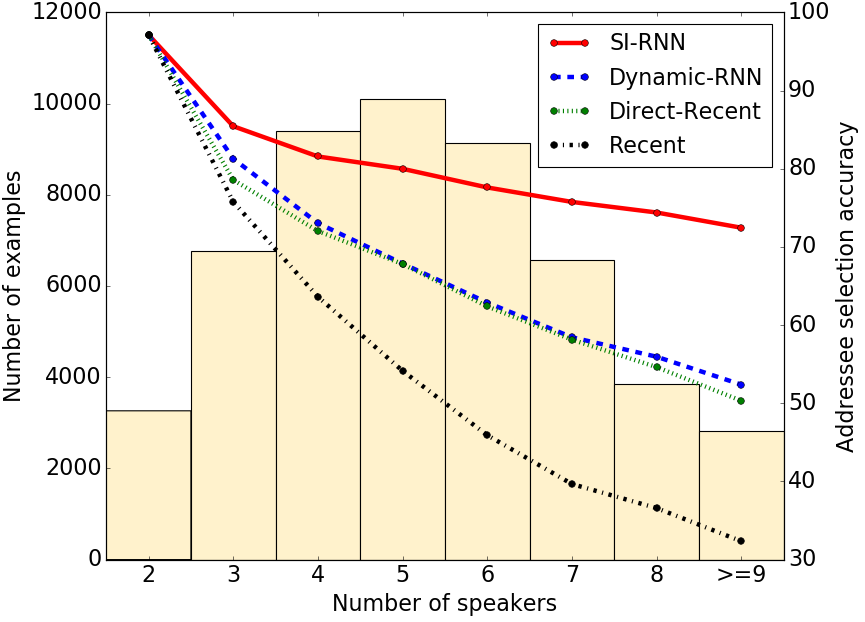}
  \label{fig:context_bin}
\end{subfigure}
\begin{subfigure}{.45\textwidth}
  \centering
  \includegraphics[width=.8\textwidth]{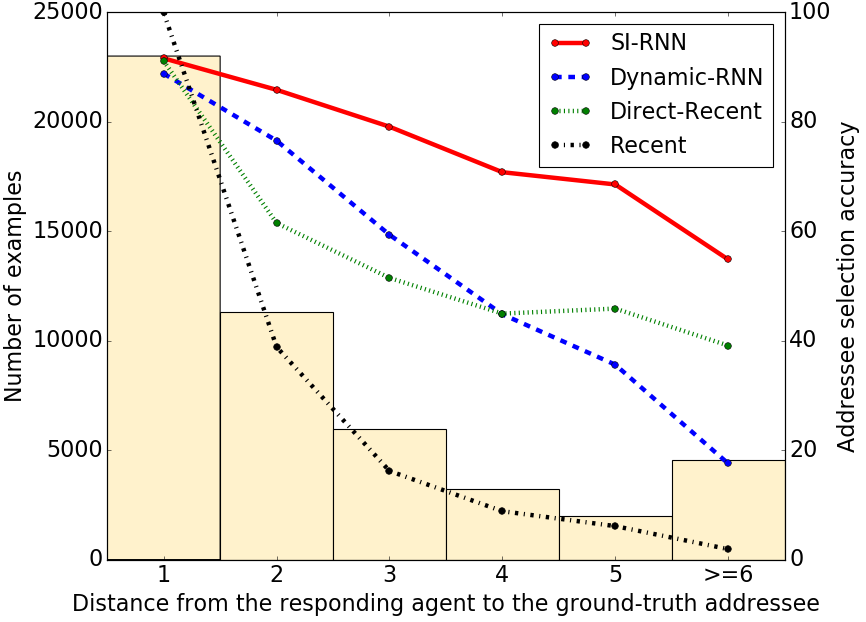}
  \label{fig:distance_bin}
\end{subfigure}
\caption{Effect of the number of speakers in the context (Upper) and the addressee distance (Lower).
Left axis: the histogram shows the number of test examples. Right axis: the curves show \texttt{ADR} accuracy on the test set.}
\label{fig:acc_bin}
\end{figure}
\noindent
\textbf{Number of Speakers.}
Numerous speakers create complex dialogs and increased candidate addressee, thus the task becomes more challenging. 
In Figure \ref{fig:acc_bin} (Upper), we investigate how \texttt{ADR} accuracy changes with the number of speakers in the context of length 15, corresponding to the rows with T=15 in Table \ref{tab:ubuntu_result}.
\recentTFIDF{} always chooses the most recent speaker and the accuracy drops dramatically as the number of speakers increases.
\directRecentTFIDF{} shows better performance, and \dynamicRNN is marginally better. 
SI-RNN is much more robust and remains above 70\% accuracy across all bins.
The advantage is more obvious for bins with more speakers.
\\\textbf{Addressing Distance.}
Addressing distance is the time difference from the responding speaker to the ground-truth addressee.
As the histogram in Figure \ref{fig:acc_bin} (Lower) shows, while the majority of responses target the most recent speaker, many responses go back five or more time steps.
It is important to note that for those distant responses, \dynamicRNN{} sees a clear performance decrease, even worse than \directRecentTFIDF.
In contrast, SI-RNN handles distant responses much more accurately.

\urlstyle{same}
\begin{table*}[t!]
\small
\footnotesize
\begin{subtable}{1\textwidth}
\begin{tabular}{c|c|c|p{12.cm}}
& Sender      & Addressee  & Utterance  \\ \hline
1 & codepython  & wafflejock & thanks \\
1 & wafflejock      & codepython & yup np \\
2 & wafflejock & theoletom & you can use sudo apt-get install packagename -- reinstall, to have apt-get install reinstall some package/metapackage and redo the configuration for the program as well \\
3 & codepython  & - & i installed ubuntu on a separate external drive. now when i boot into mac, the external drive does not show up as bootable. the blue light is on. any ideas? \\
4 & Guest54977      & - & hello there. wondering to anyone who knows, where an ubuntu backup can be retrieved from. \\
2 & theoletom & wafflejock & it's not a program. it's a desktop environment. \\
4 & Guest54977      & - & did some searching on my system and googling, but couldn't find an answer \\
2 & theoletom       & - & be a trace of it left yet there still is. \\
2 & theoletom       & - & i think i might just need a fresh install of ubuntu. if there isn't a way to revert to default settings \\
5 & releaf  & - & what's your opinion on a \$500 laptop that will be a dedicated ubuntu machine? \\
5 & releaf  & - & are any of the pre-loaded ones good deals? \\
5 & releaf  & - & if not, are there any laptops that are known for being oem-heavy or otherwise ubuntu friendly? \\
3 & codepython  & - & my usb stick shows up as bootable (efi) when i boot my mac. but not my external hard drive on which i just installed ubuntu. how do i make it bootable from mac hardware? \\
3 & Jordan\_U        & codepython & did you install ubuntu to this external drive from a different machine? \\
5 & Umeaboy & releaf & what country you from? \\
5 & wafflejock &  & \\ \Xhline{4\arrayrulewidth}
\multicolumn{2}{c|}{Model Prediction} & Addressee & Response  \\ \hline
\multicolumn{2}{c|}{Direct-Recent+TF-IDF} & theoletom & ubuntu install fresh \\
\multicolumn{2}{c|}{Dynamic-RNN} & codepython &  no prime is the replacement \\
\multicolumn{2}{c|}{SI-RNN}  & $\star$ releaf & $\star$ there are a few ubuntu dedicated laptop providers like umeaboy is asking depends on where you are \\
\end{tabular}
\caption{
SI-RNN chooses to engage in a new sub-conversation by suggesting a solution to ``releaf" about Ubuntu dedicated laptops.}
\end{subtable}

\begin{subtable}{1\textwidth}
\begin{tabular}{c|c|c|p{11.2cm}}
& Sender      & Addressee  & Utterance  \\ \hline
1 & VeryBewitching  & nicomachus & anything i should be concerned about before i do it? \\
1 & nicomachus      & VeryBewitching & always back up before partitioning. \\
1 & VeryBewitching  & nicomachus & i would have assumed that, i was wondering more if this is something that tends to be touch and go, want to know if i should put coffee on : ) \\
2 & TechMonger      & - & it was hybernating. i can ping it now \\
2 & TechMonger      & - & why does my router pick up disconnected devices when i reset my device list? or how \\
2 & Ionic   & - & because the dhcp refresh interval hasn't passed yet? \\
2 & TechMonger      & - & so dhcp refresh is different than device list refresh? \\
2 & D33p    & TechMonger & what an enlightenment @techmonger : ) \\
2 & BuzzardBuzz     & - & dhcp refresh for all clients is needed when you change your subnet ip \\
2 & BuzzardBuzz     & - & if you want them to work together \\
2 & Ionic   & BuzzardBuzz & uhm, no. \\
2 & chingao & TechMonger & nicomachus asked this way at the beginning: is the machine that you 're trying to ping turned on? \\
1 & nicomachus & & \\
\Xhline{4\arrayrulewidth}
\multicolumn{2}{c|}{Model Prediction} & Addressee & Response  \\ \hline
\multicolumn{2}{c|}{Direct-Recent+TF-IDF} & $\star$ VeryBewitching & i have tried with this program y-ppa manager, yet still doesn't work.\\
\multicolumn{2}{c|}{Dynamic-RNN} & chingao & install the package ``linux-generic'', that will install the kernel and the headers if they are not installed \\
\multicolumn{2}{c|}{SI-RNN}  & $\star$ VeryBewitching &  $\star$ if it's the last partition on the disk, it won't take long. if gparted has to copy data to move another partition too, it can take a couple hours. \\
\end{tabular}
\caption{
SI-RNN remembers the distant sub-conversation 1 and responds to ``VeryBewitching" with a detailed answer.
}
\end{subtable}
\caption{Case Study. $\star$ denotes the ground-truth. Sub-conversations are coded with different numbers for the purpose of analysis (sub-conversation labels are not available during training or testing).}
\label{tb:example}
\end{table*}

\noindent
\textbf{Case Study.}
Examples in Table \ref{tb:example} show how SI-RNN can handle complex multi-party conversations by selecting from 10 candidate responses.
In both examples, the responding speakers participate in two or more concurrent sub-conversations with other speakers.

Example (a) demonstrates the ability of SI-RNN to engage in a new sub-conversation.
The responding speaker ``wafflejock" is originally involved in two sub-conversations: the sub-conversation 1 with ``codepython", and the ubuntu installation issue with ``theoletom".
While it is reasonable to address ``codepython" and ``theoletom", the responses from other baselines are not helpful to solve corresponding issues.
\textsc{TF-IDF} prefers the response with the ``install" key-word, yet the response is repetitive and not helpful.
\dynamicRNN{} selects an irrelevant response to ``codepython".
SI-RNN chooses to engage in a new sub-conversation by suggesting a solution to ``releaf" about Ubuntu dedicated laptops.

Example (b) shows the advantage of SI-RNN in responding to a distant message.
The responding speaker ``nicomachus" is actively engaged with ``VeryBewitching" in the sub-conversation 1 and is also loosely involved in the sub-conversation 2: ``chingao" mentions ``nicomachus" in the most recent utterance.
SI-RNN remembers the distant sub-conversation 1 and responds to ``VeryBewitching" with a detailed answer.
\directRecentTFIDF{} selects the ground-truth addressee because ``VeryBewitching" talks to ``nicomachus", but the response is not helpful.
\dynamicRNN{} is biased to the recent speaker ``chingao", yet the response is not relevant.

\section{Conclusion}
SI-RNN jointly models {\em who} says {\em what} to {\em whom} by updating speaker embeddings in a role-sensitive way.
It provides state-of-the-art addressee and response selection, which can instantly help retrieval-based dialog systems.
In the future, we also consider using SI-RNN to extract sub-conversations in the unlabeled conversation corpus and provide a large-scale disentangled multi-party conversation data set.

\section{Acknowledgements}
We thank the members of the UMichigan-IBM Sapphire Project and all the reviewers for their helpful feedback.
This material is based in part upon work supported by IBM under contract 4915012629.
Any opinions, findings, conclusions or recommendations expressed above are those of the authors and do not necessarily reflect the views of IBM.

\bibliographystyle{aaai}
\bibliography{aaai}

\end{document}